\documentstyle[colap]{article}

\title{\vspace{-0.5in}Concept clustering and knowledge integration from a children's dictionary}
\author{Caroline Barri\`{e}re and Fred Popowich \\
School of Computing Science, Simon Fraser University \\ Burnaby, BC, Canada, V5A 1S6 \\barriere,popowich@cs.sfu.ca}

\begin{document}

\maketitle
\vspace{-0.5in}
\begin{abstract}
Knowledge structures called Concept Clustering Knowledge Graphs (CCKGs) are introduced along with a process for their construction from a machine readable dictionary.
CCKGs contain multiple concepts interrelated through multiple semantic relations together forming a semantic cluster represented by a conceptual graph.  
The knowledge acquisition is performed on a children's first dictionary.
The concepts involved are general and typical of a daily life conversation.
A collection of conceptual clusters together can form the basis of a lexical knowledge base, where each CCKG contains a limited number of highly connected words giving useful information about a particular domain or situation.
\end{abstract}

\section{Introduction}

When constructing a Lexical Knowledge Base (LKB) useful for Natural Language Processing, the source of information from which knowledge is acquired and the structuring of this information within the LKB are two key issues.  Machine Readable Dictionaries (
MRDs) are a good source of lexical information and have been shown to be applicable to the task of LKB construction \cite{Dolan_et_al93,Calzolari92,Copestake90,Wilks_et_al89,Byrd_et_al87}.  Often though, a localist approach is adopted whereby the words ar
e kept in alphabetical order with some representation of their definitions in the form of a template or feature structure.  
Efforts in finding connections between words is seen in work on automatic extraction of semantic relations from MRDs  \cite{Ahlswede_et_Evens88,Alshawi89,Montemagni_et_Vanderwende92}.
Additionally, efforts in finding words that are close semantically is seen by the current interest in statistical techniques for word clustering, looking at co-occurrences of words in text corpora or dictionaries \cite{Church_et_Hanks89,Wilks_et_al89,Brow
n_et_al92,Pereira_et_al95}.

Inspired by research in the areas of semantic relations, semantic distance, concept clustering, and using Conceptual Graphs \cite{Sowa84} as our knowledge representation, we introduce Concept Clustering Knowledge Graphs (CCKGs).
Each CCKG will start as a Conceptual Graph representation of a trigger word and will expand following a search algorithm to incorporate related words and form a Concept Cluster.  The concept cluster in itself is interesting for tasks such as word disambig
uation, but the CCKG will give more to that cluster.  It will give the relations between the words, making the graph in some aspects similar to a script \cite{Schank_et_Abelson75}.  However, a CCKG is generated automatically and does not rely on primitive
s but on an unlimited number of concepts, showing objects, persons, and actions interacting with each other.  This interaction will be set within a particular domain, and the trigger word should be a key word of the domain to represent.  If that process w
ould be done for the whole dictionary, we would obtain an LKB divided into multiple clusters of words, each represented by a CCKG.  Then during text processing for example, a portion of text could be analyzed using the appropriate CCKG to find implicit re
lations and help understanding the text.

Our source of knowledge is the American Heritage First Dictionary\footnote{Copyright \copyright1994 by Houghton Mifflin Company.  Reproduced by permission from THE AMERICAN HERITAGE FIRST DICTIONARY.} which contains 1800 entries and is designed for childr
en of age six to eight.
It is made for young people learning the structure and the basic vocabulary of their language.
In comparison, an adult's dictionary is more of a reference tool which assumes knowledge of a large basic vocabulary, 
while a learner's dictionary assumes a limited vocabulary but still some
very sophisticated concepts.
Using a children's dictionary allows us to restrict our vocabulary, 
but still work on general knowledge about day to day concepts and actions.

In the following sections, we first present the transformation steps from the definitions into conceptual graphs, then we elaborate on the integration process, and finally, we close with a discussion.

\section{Transforming Definitions}

Our definitions may contain up to three general types of information, as shown in the examples in Figure~\ref{ex1}.  
{\small
\begin{itemize}
   \item {\bf description: } This contains genus/differentia information.
Such information is frequently used for noun taxonomy construction \cite{Byrd_et_al87,Klavans_et_al90,Barriere_et_Popowich_EURALEX96}.
   \item {\bf general knowledge or usage: } This gives information useful in daily life, like how to use an object, what it is made of, what it looks like, etc.
   \item {\bf specific example: } This presents a typical situation using the word defined and it involves specific persons and actions.
\end{itemize}
}
\begin{figure}[htbp]
\centering{
\framebox[3.0in][c]{ \parbox[t]{2.8in} {\scriptsize
{\bf Cereal} is a kind of food. [description] \\
Many {\bf cereals} are made from corn, wheat, or rice. [usage] \\
Most people eat {\bf cereal} with milk in a bowl. [usage] \\ \\
{\bf Ash} is what is left after something burns. [usage] \\
It is a soft gray powder. [description] \\
Ray watched his father clean the {\bf ashes} out of the fireplace. [example]
}
}
}
\caption{Example of definitions\label{ex1}}
\end{figure}

The information given by the description and general knowledge will be used to perform the knowledge integration proposed in section~\ref{integ}.
The specific examples are excluded as they tend to involve specific concepts not always deeply related to the word defined.  

Our processing of the definitions results in the construction of a special type of conceptual graph which we call a {\em temporary graph}.  The set of relations used in temporary graphs come from three sources.  Table~\ref{ex_rel} shows some examples for 
each type.

{\small
\begin{enumerate}
   \item the set of closed class words, ex: {\em of, to, in, and};
   \item relations extracted via defining formulas ex: {\em part-of, made-of, instrument};  defining formulas correspond to phrasal patterns that occur often through the dictionary suggesting particular semantic relations (ex. A is a part of B) \cite{Ahls
wede_et_Evens88,Dolan_et_al93}. 
   \item the relations that are extracted from the syntactic structure of a sentence, ex: {\em subject, object, goal, attribute, modifier}.
\end{enumerate}
}

As some relations are defined using the closed class words, and many of those words are ambiguous, the resulting graph will itself be ambiguous.  This is the main reason for calling our graphs {\em temporary} as we assume a conceptual graph, the ultimate 
goal of our translation process, should contain a restricted set of well-defined and non-ambiguous semantic relations.  For example, {\em by} can be a relation of manner ({\em by chewing}), time ({\em by noon}) or place ({\em by the door}).  By keeping th
e preposition itself within the temporary graph, we delay the ambiguity resolution process until we have gathered more information and we even hopefully avoid the decision process as the ambiguity might later be resolved by the integration process itself.

\begin{table}[htbp]     
\centering{\scriptsize
\begin{tabular}{|l|l|} \hline
1. closed class words & temporary graph \\ \hline
np:np[A],prep[B],np[C] & [A]-$>$(B)-$>$[C] \\
apple on the table & [apple]-$>$(on)-$>$[table] \\ & \\ \hline
2. defining formulas & temporary graph \\ \hline
A is used to B & [B]-$>$(instrument)-$>$[A] \\
A is a part of B  & [A]-$>$(part-of)-$>$[B] \\
A is a place where B & [B]-$>$(loc)-$>$[A] \\ & \\ \hline
3. syntactic pattern & temporary graph \\ \hline
s:np[A],vp[B]   & [B]-$>$(agent)-$>$[A] \\
John eats  & [eat]-$>$(agent)-$>$[John] \\ 
          &                            \\
 vp:vp[A],inf\_vp[B] & [A]-$>$(goal)-$>$[B] \\ 
eat to grow & [eat]-$>$(goal)-$>$[grow] \\ \hline
\end{tabular}
}
\caption{Examples of relations found in sentences and their corresponding temporary graphs \label{ex_rel}}
\end{table}

\section{Knowledge Integration} \label{integ}

This section describes how given a trigger word, we perform a series of forward and backward searches in the dictionary to build a CCKG containing useful information pertaining to the trigger word and to closely related words.  The primary building blocks
 for the CCKG are the temporary graphs built from the dictionary definitions of those words using our transformation process described in the previous section.  Those temporary graphs express similar or related ideas in different ways and with different l
evels of detail.  As we will try to put all this information together into one large graph, we must first find what information the various temporary graphs have in common and then join them around this common knowledge.

To help us build this CCKG and perform our integration process, we assume two main knowledge structures are available, a concept hierarchy and a relation hierarchy, and we assume the existance of some graph operations.  The concept hierarchy concentrates 
on nouns and verbs as they account for three quarters of the dictionary definitions.  It has been constructed automatically according to the techniques described in \cite{Barriere_et_Popowich_EURALEX96}.  The relation hierarchy was constructed manually.  
A rich hierarchical structure between the set of relations is essential to the graph matching operations we use for the integration phase.

As we are using the conceptual graph formalism to represent our definitions, we can use the graph matching operations defined in \cite{Sowa84}.  The two operations we will need are the Maximal Common Subgraph algorithm and the Maximal Join algorithm.  

\subsection{Maximal Common Subgraph}

The maximal common subgraph between two graphs consists of finding a subgraph of the first graph that is isomorphic to a subgraph of the second graph.  In our case, we cannot often expect to find two graphs that contain an identical subgraph with the exac
t same relations and concepts.  Ideas can be expressed in many ways and we therefore need a more relaxed matching schema.  We describe a few elements of this ``relaxation'' process and illustrate them by an example in Figure~\ref{ex2}.

\begin{figure}[htbp]
\centering{
\framebox[3.0in][c]{ \parbox[t]{2.8in} {\tiny
\begin{tabbing}

   (1) John makes a nice drawing on a piece of paper with the pen. \\

   {}[make]\=-$>$(sub)-$>$[John] \\
          \>-$>$(obj)-$>$[drawing]-$>$(att)-$>$[nice] \\
          \>-$>$(on)-$>$[piece]-$>$(of)-$>$[paper] \\
          \>-$>$(with)-$>$[pen] \\ \\

   (2) John uses the big crayon to draw rapidly on the paper. \\
   {}[draw]\=-$>$(sub)-$>$[John] \\
         \>-$>$(on)-$>$[paper] \\
         \>-$>$(instrument)-$>$[crayon] \\
         \>-$>$(manner)-$>$[rapidly] \\ \\

   MAXIMAL COMMON SUBGRAPH: \\
   {}[make(draw)]\=-$>$(sub)-$>$[John] \\
          \>-$>$(obj)-$>$[drawing] \\
          \>-$>$(on)-$>$[piece]-$>$(of)-$>$[paper] \\
          \>-$>$(instrument)-$>$[label-1] \\

\end{tabbing}
\begin{tabular}{|l|l|l|}\hline
  Relaxation method & graph1 & graph2  \\ \hline
  Semantic distance & pen    & crayon \\
   Relation subsumption & with & instrument \\
   Predictable meaning shift & drawing & draw \\
   Relation transitivity & piece of paper & paper \\ \hline
\end{tabular}
\\
\begin{tabbing}
   MAXIMAL JOIN: \\
   {}[make(draw)]\=-$>$(sub)-$>$[John] \\
          \>-$>$(obj)-$>$[drawing]-$>$(att)-$>$[nice] \\
          \>-$>$(on)-$>$[piece]-$>$(of)-$>$[paper] \\
          \>-$>$(instrument)-$>$[label-1] \\ 
          \>-$>$(manner)-$>$[rapidly] \\ \\

\end{tabbing}

}
}
}
\caption{Example of ``relaxed'' maximal common subgraph and maximal join algorithms \label{ex2}}
\end{figure}

\paragraph{Semantic distance between concepts.}  In the maximal common subgraph algorithm proposed by \cite{Sowa84}, two concepts (C1,C2) could be matched if one subsumed the other in the concept hierarchy.  We can relax that criteria to match two concept
s when a third concept C which subsumes C1 and C2 has a high enough degree of informativeness \cite{Resnik95a}.  The concept hierarchy can be useful in many cases, but it is generated from the dictionary and might not be complete enough to find all simila
r concepts.

   In the example of Figure~\ref{ex2}, when using the concept hierarchy to establish the similarity between {\em pen} and {\em crayon}, we find that one is a subclass of {\em tool} and the other of {\em wax}, both then are subsumed by the general concept 
{\em something}.  We have reached the root of the noun tree in the concept hierarchy and this would give a similarity of 0 based on the informativeness notion.

We extend the subsumption notion to the graphs.  Instead of finding a concept that subsumes two concepts, we will try finding a common subgraph that subsumes the graph representation of both concepts.
In our example, {\em pen} and {\em crayon} have a common subgraph [write]-$>$(inst)-$>$[].  The notion of semantic distance can be seen as the informativeness of the subsuming graph.  The resulting maximal common subgraph as shown in Figure~\ref{ex2} cont
ains the concept label-1.  This label is associated to a covert category as presented in \cite{Barriere_et_Popowich_EURALEX96}.  We can update the concept hierarchy and add this label-1 as a subclass of {\em something} and a superclass of {\em pen} and {\
em crayon}.  It expresses a concept of ``writing instrument''.

\paragraph{Relation subsumption.} Since we have a relation hierarchy in addition to our concept hierarchy, we can similarly use subsumption to match two relations.
In Figure~\ref{ex2}, {\em with} is subsumed by {\em instrument}, and by mapping them, we disambiguate {\em with} from corresponding to another semantic relation, such as {\em possession} or {\em accompaniment}.  This is a case where an ambiguous prepositi
on left in the temporary graph is resolved by the integration process.

\paragraph{Predictable meaning shift.}  A set of lexical implication rules were developed by \cite{Ostler_et_Atkins92} for relating word senses.  Based on them, we are developing a set of graph matching rules.  Figure~\ref{ex2} exemplifies one of them whe
re two graphs containing the same word (or morphologically related), here {\em draw} and {\em drawing}, used as different parts of speech can be related.

\paragraph{Relation transitivity.} Some relations, like {\em part-of}, {\em in}, {\em from} can be transitive.  For example, we can map a graph that contains a concept A in a certain relation to concept B onto another graph where concept A is in the same 
relation with a {\em part} or a {\em piece} of B as exemplified in Figure~\ref{ex2}.  Transitivity in relations is in itself a challenging area of study \cite{Cruse86} and we have only begun to explore it.

\subsection{Maximal Join}

The basic operation for the integration of temporary graphs is the maximal join operation where a union of two graphs is formed around their maximal common subgraph using the most specific concepts of each.  We just saw how to relax the maximal common sub
graph operation and we will perform the join around that ``relaxed'' subgraph.  Figure~\ref{ex2} shows the result of the maximal join.  The join operation allows us to bring new concepts into a graph by finding relations with existing concepts, as well as
 bringing new relations between existing concepts.

\subsection{Integration process}

Given the concept hierarchy, relation hierarchy and graph matching operations, 
we now describe the two major steps required to integrate all the temporary graphs into a CCKG.

\paragraph{TRIGGER PHASE.}  Start with a central word, a keyword for the subject of interest that becomes the trigger word. The temporary graph built from the trigger word forms the initial CCKG.  To expand its meaning, we want to look at the important co
ncepts involved and use their respective temporary graphs to extend our initial graph.  We deem words in the definition to be important if they have a large semantic weight.  

The semantic weight of a word or its informativeness can be related to its frequency \cite{Resnik95a}.  Here, we calculate the number of occurrence of each word within the definitions of nouns and verbs in our dictionary.  The most frequent word ``a'' occ
urs 2600 times among a total of 38000 word occurrences.  Only 1\% of the words occur more than 130 times, 5\% occur more than 30 times but over 60\% occur less than 5 times.  

Ordering the dictionary words in terms of decreasing number of occurrences, the top 10\% of these words account for 75\% of word occurrences.
For our current investigation, we propose this as the division between semantically significant words, and semantically insignificant ones.
So a word from the dictionary is deemed to be semantically significant if it occurs less than 17 times.
Note that constraining the number of semantically significant words is important in limiting the exploration process for constructing the concept cluster, as we shall soon see.
   \begin{description}
       \item [Trigger forward: ] Find the semantically significant words part of the CCKG, and join their respective temporary graph to the initial CCKG.
       \item [Trigger backward: ] Find all the words in the dictionary that use the trigger word in their definition and join their respective temporary graph to the CCKG.
   \end{description}

Instead of a single trigger word, we now have a cluster of words that are related through the CCKG.  Those words form the {\em concept cluster}.

\paragraph{EXPANSION PHASE.} We try finding words in the dictionary containing many concepts identical to the ones already present in the CCKG but perhaps interacting through different relations allowing us to create additional links within the set of con
cepts present in the CCKG.  Our goal is to create a more interconnected graph rather than sprouting from a particular concept.  For this reason, we establish a graph matching threshold to decide whether we will join a new graph to the CCKG being built.  W
e set this threshold empirically: the maximal common subgraph between the CCKG and the new temporary graph must contain at least three concepts connected through two relations.

  \begin{description}
     \item  [Expansion forward: ] For each semantically significant word in the CCKG, not already part of the concept cluster, find the maximal common subgraph between its temporary graph and the CCKG.  If matching surpasses the graph matching threshold, 
perform integration (maximal join operation) and add the word in the concept cluster.  Continue forward until no changes are made.
      \item [Expansion backward: ] Find words in the dictionary whose definitions contain the semantically significant words from the concept cluster.  For each possible new word, perform the maximal common subgraph between its temporary graph and the CCK
G.  Again, if matching is over the graph matching threshold, perform integration and add the word in the concept cluster.  Continue until no changes are made.
   \end{description}

We can set a limit to the number of steps in the expansion phase to ensure its termination.  However in practice, after two or three steps forward or backward, the maximal common subgraphs between the new graphs and CCKG do not exceed the graph matching t
hreshold and thus are not added to the cluster, terminating the expansion.

\subsection{Example of integration}

\begin{figure}[htbp]
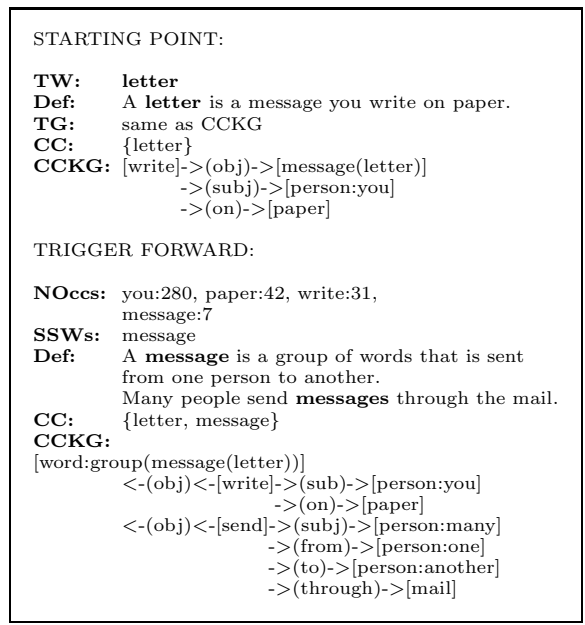

\centering{
\framebox[3.0in][c]{ \parbox[t]{2.5in} {\scriptsize
\begin{tabbing}
xxxxxxxx\= \kill
STARTING POINT: \\ \\
{}{\bf TW:} \>{\bf letter} \\
{}{\bf Def:} \>A {\bf letter} is a message you write on paper. \\
{}{\bf TG:} \>same as CCKG \\
{}{\bf CC:} \>\{letter\} \\
{}{\bf CCKG:} \>[write]\=-$>$(obj)-$>$[message(letter)] \\
           \>       \>-$>$(subj)-$>$[person:you] \\
           \>       \>-$>$(on)-$>$[paper] \\ \\
TRIGGER FORWARD: \\ \\
{}{\bf NOccs:}\>you:280, paper:42, write:31, \\
{}                       \> message:7 \\
{}{\bf SSWs:} \>message \\
{}{\bf Def:} \> A {\bf message} is a group of words that is sent\\
{}          \> from one person to another. \\  
{}          \> Many people send {\bf messages} through the mail. \\
{}{\bf CC:} \> \{letter, message\} \\
{}{\bf CCKG: } \\
{}[word:group(message(letter))] \\
{}           \>$<$-(obj)$<$-[write]\=-$>$(sub)-$>$[person:you] \\
{}            \>                   \>-$>$(on)-$>$[paper]\\
{}            \>$<$-(obj)$<$-[send]\=-$>$(subj)-$>$[person:many]\\
{}            \>               \>-$>$(from)-$>$[person:one] \\
{}            \>               \>-$>$(to)-$>$[person:another] \\
{}            \>               \>-$>$(through)-$>$[mail]
\end{tabbing}
}
}
}
\caption{Trigger forward from {\em letter}. \label{ex}}
\end{figure}

Figure~\ref{ex} shows the starting point of an integration process with the trigger word (TW) {\em letter}, its definition, its temporary graph (TG), the concept cluster (CC) containing only the trigger word, and the CCKG being the same as the temporary g
raph.  Then we show the trigger forward phase.  The number of occurences (NOcc) of each word present in the definition of {\em letter} is given.  Using the criteria described in the previous section, only the word {\em message} is a semantically significa
nt word (SSW).  We then see the definition of message, the new concept cluster and the resulting CCKG.

The trigger backward phase, would incorporate the temporary graphs for {\it address, mail, post office} and {\it stamp}.  The expansion forward phase would further add the temporary graphs for the semantically significant words: {\it \{send, package\}} du
ring the first step and then would terminate with the second step as no more semantically significant words not yet explored have a maximal common subgraph with the CCKG that exceeds the graph matching threshold.  The expansion backward would finally add 
the TGs for {\it card} and {\it note}, again terminating after two steps.  

The resulting cluster is: \{letter, message, address, mail, post office, stamp, send, package, card, note\}.  The resulting CCKG shows the interaction between those concepts which summarizes general knowledge about how we use those concepts together in a 
daily conversation: we go to the post office to mail letters, or packages; we write letters, notes and card to send to people through the mail, etc.  Having such clusters and such knowledge of the relationship between words as part of our lexical knowledg
e base can be useful to understand or even generate a text containing the concepts involved in the cluster.

\section{Discussion}
Through this paper, we showed the multiple steps leading us to the building of Concept Clustering Knowledge Graphs (CCKGs).  Those knowledge structures are built within the Lexical Knowledge Base (LKB), integrating multiple parts of the LKB around a parti
cular concept to form a cluster and express the multiple relations among the words in that cluster.  The CCKGs could be either permanent or temporary structures depending on the application using the LKB.  For example, for a text understanding task, we ca
n build before hand the CCKGs corresponding to one or multiple keywords from the text.  Once built, the CCKGs will help us in our comprehension and disambiguation of the text.

By using the American Heritage First Dictionary as our source of lexical information, we were able to restrict our vocabulary to result in a project of reasonable size, dealing with general knowledge about day to day concepts and actions.  The ideas explo
red using this dictionary can be extended to other dictionaries as well, but the task might become more complex as the definitions in adult's dictionaries are not as clear and usage oriented.  In fact, an LKB built from a children's dictionary could be se
en as a starting point from which we could extend our acquisition of knowledge using text corpora or other dictionaries.  Certainly, if we envisage applications trying to understand children's stories or help in child education, a corpora of texts for chi
ldren would be a good source of information to extend our LKB.

The graph operations (maximal common subgraph and maximal join) defined on conceptual graphs, and adapted here, play an important role in our integration process toward a final CCKG.
Graph matching was also suggested as an alternative to taxonomic search when trying to establish semantic similarity between concepts.  As well, by putting a threshold on the graph matching process, we were able to limit the expansion of our clustering, a
s we can decide and justify 
the incorporation of a new concept into a particular cluster.

Many aspects of the concept clustering and knowledge integration processes have already been implemented and it will soon be possible to test the techniques on different trigger words using different thresholds to see how they effect the quality of the cl
usters.

Clustering is often seen as a statistical operation that puts together words ``somehow'' related.  Here, we give a meaning to their clustering, we find and show the connections between concepts, and by doing so, we build more than a cluster of words. We b
uild a knowledge graph where the concepts interact with each other giving important implicit information that will be useful for Natural Language Processing tasks.

\section{Acknowledgments}
The authors would like to thank the anonymous referees for their comments and suggestions.  This research was supported by the Institute for Robotics and Intelligent Systems.  The authors also want to thank Petr Kubon for his many comments on the paper.


\begin{thebibliography}{}

\bibitem[\protect\citename{Ahlswede and Evens}1988]{Ahlswede_et_Evens88}
T.~Ahlswede and M.~Evens.
\newblock 1988.
\newblock Generating a relational lexicon from a machine-readable dictionary.
\newblock {\em International Journal of Lexicography}, 1(3):214--237.

\bibitem[\protect\citename{Alshawi}1989]{Alshawi89}
H.~Alshawi.
\newblock 1989.
\newblock Analysing the dictionary definitions.
\newblock In Bran Boguraev and Ted Briscoe, editors, {\em Computational
  Lexicography for Natural Language Processing}, chapter~7, pages 153--170.
  Longman Group UK Limited.

\bibitem[\protect\citename{Barri\`{e}re and
  Popowich}1996]{Barriere_et_Popowich_EURALEX96}
Caroline Barri\`{e}re and Frederick Popowich.
\newblock 1996.
\newblock Building a noun taxonomy from a children's dictionary.
\newblock In {\em Euralex'96}.
\newblock To be presented at Euralex'96, G\^oteborg, Sweden, August 96.

\bibitem[\protect\citename{Brown \bgroup et al.\egroup }1992]{Brown_et_al92}
P.~Brown, V.J.~Della Pietra, P.V. deSouza, J.C. Lai, and R.L. Mercer.
\newblock 1992.
\newblock Class-based n-gram models of natural language.
\newblock {\em Computational Linguistics}, 18(4):467--480.

\bibitem[\protect\citename{Byrd \bgroup et al.\egroup }1987]{Byrd_et_al87}
R.J. Byrd, N.~Calzolari, M.~Chodorow, J.~Klavans, M.~Neff, and O.~Rizk.
\newblock 1987.
\newblock Tools and methods for computational lexicology.
\newblock {\em Computational Linguistics}, 13(3-4):219--240.

\bibitem[\protect\citename{Calzolari}1992]{Calzolari92}
N.~Calzolari.
\newblock 1992.
\newblock Acquiring and representing semantic information in a lexical
  knowledge base.
\newblock In J.~Pustejovsky and S.~Bergler, editors, {\em Lexical Semantics and
  Knowledge Representation : First SIGLEX Workshop}, chapter~16, pages
  235--244. Springer-Verlag.

\bibitem[\protect\citename{Church and Hanks}1989]{Church_et_Hanks89}
K.~Church and P.~Hanks.
\newblock 1989.
\newblock Word association norms, mutual information and lexicography.
\newblock In {\em Proceedings of the 27th Annual meeting of the Association for
  Computational Linguistics}, pages 76--83, Vancouver, BC.

\bibitem[\protect\citename{Copestake}1990]{Copestake90}
A.A. Copestake.
\newblock 1990.
\newblock An approach to building the hierarchical element of a lexical
  knowledge base from a machine readable dictionary.
\newblock In {\em Proceedings of the Workshop on Inheritance in Natural
  Language Processing, Tilburg}.

\bibitem[\protect\citename{Cruse}1986]{Cruse86}
D.A. Cruse.
\newblock 1986.
\newblock {\em Lexical Semantics}.
\newblock Cambridge University Press.

\bibitem[\protect\citename{Dolan \bgroup et al.\egroup }1993]{Dolan_et_al93}
W.~Dolan, L.~Vanderwende, and S.~D. Richardson.
\newblock 1993.
\newblock Automatically deriving structured knowledge bases from on-line
  dictionaries.
\newblock In {\em The First Conference of the Pacific Association for
  Computational Linguistics}, pages 5--14, Harbour Center, Campus of SFU,
  Vancouver, April.

\bibitem[\protect\citename{Klavans \bgroup et al.\egroup
  }1990]{Klavans_et_al90}
Judith Klavans, M.~S. Chodorow, and N.~Wacholder.
\newblock 1990.
\newblock From dictionary to knowledge base via taxonomy.
\newblock In {\em Proceedings of the 6th Annual Conference of the UW Centre for
  the New OED: Electronic Text Research}, pages 110--132.

\bibitem[\protect\citename{Montemagni and
  Vanderwende}1992]{Montemagni_et_Vanderwende92}
S.~Montemagni and L.~Vanderwende.
\newblock 1992.
\newblock Structural patterns vs. string patterns for extracting semantic
  information from dictionaries.
\newblock In {\em Proc.\ of the 14$^{~th}$ COLING}, pages 546--552, Nantes,
  France.

\bibitem[\protect\citename{Ostler and Atkins}1992]{Ostler_et_Atkins92}
N.~Ostler and B.T.S. Atkins.
\newblock 1992.
\newblock Predictable meaning shift: Some linguistic properties of lexical
  implication rules.
\newblock In J.~Pustejovsky and S.~Bergler, editors, {\em Lexical Semantics and
  Knowledge Representation : First SIGLEX Workshop}, chapter~7, pages 87--100.
  Springer-Verlag.

\bibitem[\protect\citename{Pereira \bgroup et al.\egroup
  }1995]{Pereira_et_al95}
F.~Pereira, N.~Tishby, and L.~Lee.
\newblock 1995.
\newblock Distributional clustering of english words.
\newblock In {\em Proc.\ of the 33$^{~th}$ ACL}, Cambridge,MA.

\bibitem[\protect\citename{Resnik}1995]{Resnik95a}
Philip Resnik.
\newblock 1995.
\newblock Using information content to evaluate semantic similarity in a
  taxonomy.
\newblock In {\em Proc.\ of the 14$^{~th}$ IJCAI}, volume~1, pages 448--453,
  Montreal, Canada.

\bibitem[\protect\citename{Schank and Abelson}1975]{Schank_et_Abelson75}
R.~Schank and R.~Abelson.
\newblock 1975.
\newblock Scripts, plans and knowledge.
\newblock In {\em Advance papers 4th Intl. Joint Conf. Artificial
  Intelligence}.

\bibitem[\protect\citename{Sowa}1984]{Sowa84}
J.~Sowa.
\newblock 1984.
\newblock {\em Conceptual Structures in Mind and Machines}.
\newblock Addison-Wesley.

\bibitem[\protect\citename{Wilks \bgroup et al.\egroup }1989]{Wilks_et_al89}
Y.~Wilks, D.~Fass, G-M Guo, J.~McDonald, T.~Plate, and B.~Slator.
\newblock 1989.
\newblock A tractable machine dictionary as a resource for computational
  semantics.
\newblock In Bran Boguraev and Ted Briscoe, editors, {\em Computational
  Lexicography for Natural Language Processing}, chapter~9, pages 193--231.
  Longman Group UK Limited.

\end{thebibliography}
 \end{document}